\documentclass[10pt]{article}

\usepackage{arxiv_preprint}

\newcommand{\method}{SM4RT}
\newcommand{\methodfull}{Structured Motion 4D Reconstruction}

\newcommand{\duster}{DUSt3R\xspace}
\newcommand{\fig}[1]{Fig.~\ref{fig:#1}}
\newcommand{\imw}[2]{\includegraphics[width=#2\linewidth]{#1.png}}
\newcommand{\tb}[3]{\setlength{\tabcolsep}{#2mm}\begin{tabular}{#1}#3\end{tabular}}

\graphicspath{{figures/}}

\title{\method{}: Cascaded Feed-forward Model for 4D Reconstruction}
\shorttitle{\method{}: Cascaded Feed-forward 4D Reconstruction}

\author{
  Pang-Chi Lo\inst{1}
  \qquad
  Jhao-Syun Lai\inst{2}
  \qquad
  Tsung-Wei Ke\inst{2}
}

\affiliations{
  \inst{1}\,Carnegie Mellon University
  \qquad
  \inst{2}\,National Taiwan University
}

\abstractresources{
  \textbf{Project page}\enspace
  \href{https://sm4rt-eailab.github.io/}{sm4rt-eailab.github.io}
}

\begin{document}

\twocolumn[{%
  \maketitle
\noindent\begin{minipage}{\textwidth}
  \centering
  \includegraphics[
    width=\textwidth,
    trim=10pt 0 10pt 0,
    clip
  ]{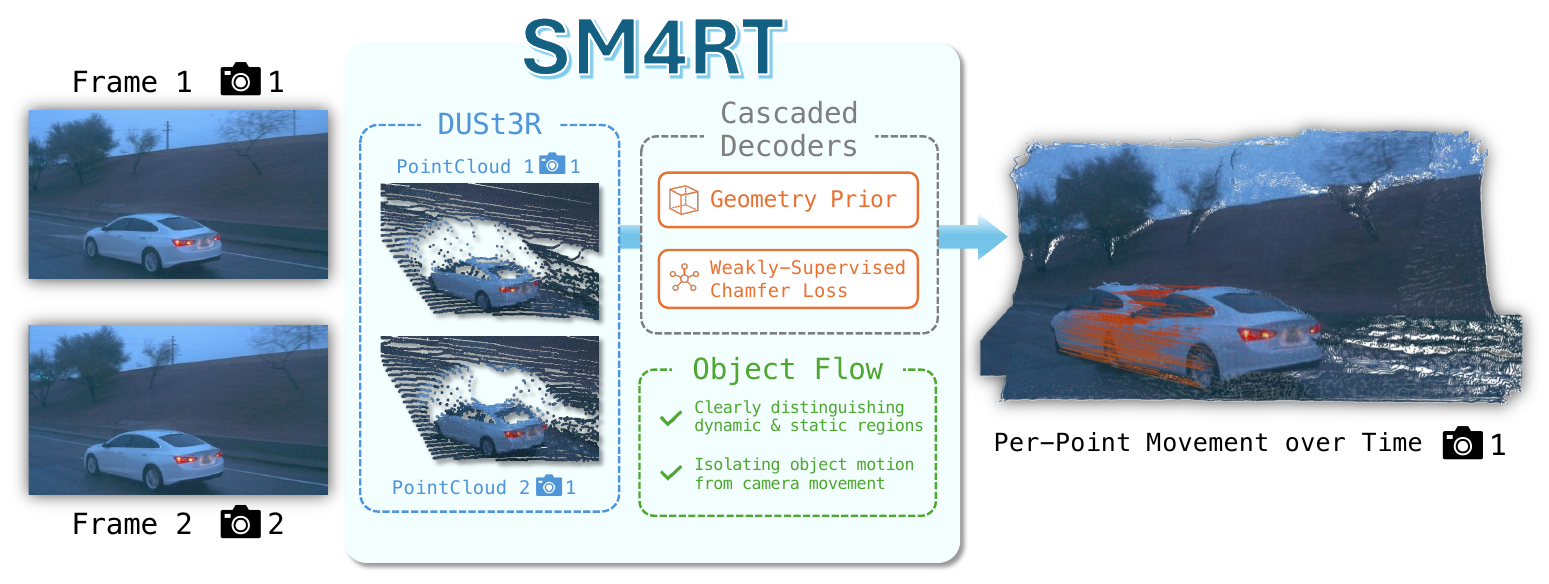}
  \vspace{4pt}
  \captionsetup{width=\textwidth}
  \captionof{figure}{%
    \textbf{\method{}} is a multi-task cascade feed-forward model that outputs
    a 3D point map for \emph{geometry reconstruction} and \emph{object flow}
    for motion estimation from two video frames. \method{} first reconstructs
    3D geometry and then bootstraps motion estimation from the predicted
    structure, achieving state-of-the-art results on multiple dynamic-scene
    benchmarks.
  }
  \label{fig:teaser}
\end{minipage}\par

  \vspace{0.62em}
}]
\thispagestyle{plain}

\begin{abstract}
We introduce \method{}, a multi-task cascade feed-forward model for dynamic scene reconstruction from RGB videos. While existing feed-forward approaches typically address geometry reconstruction and motion estimation separately--and often predict motions in a single coordinate frame that must account for camera movement--\method{} solves these tasks sequentially by first reconstructing 3D geometry and then bootstrapping motion estimation from the predicted structure. By expressing each pixel’s 3D motion in its corresponding camera frame, \method{} reduces ambiguity arising from camera motion. To address the sparsity of real-world motion labels, \method{} further incorporates a weakly supervised loss that regularizes unlabeled point motions using labeled point positions at the other time step. We evaluate motion estimation on three dynamic-scene datasets--Kubric, Waymo, and Point Odyssey--and \method{} achieves the best performance on most metrics compared with prior baselines. We additionally benchmark geometry reconstruction on five dynamic scene datasets and show that \method{} achieves competitive performance with state-of-the-art methods. Overall, our unified framework demonstrates that combining geometry reconstruction and motion estimation in a cascade not only avoids performance trade-offs but also enhances motion estimation through improved geometric reasoning.
\end{abstract}

\newpage
\section{Introduction}

Dynamic 3D scene reconstruction is fundamental to spatial understanding of the visual world.  The goal is to recover fine-grained scene geometry and motion from RGB videos, enabling a wide range of applications.  In robotics, it supports precise estimation of object pose, part articulation or point deformation; in computer vision, it facilitates simulating, novel-view rendering or editing dynamic scenes.  

Simultaneously solving geometry reconstruction and motion estimation from RGB videos remains a substantial challenge, largely due to the strong coupling between these two tasks.  Accurate motion estimation requires reliable 3D point locations at each time step~\cite{hadfield2011kinecting}, while geometry reconstruction depends on consistent multi-view correspondences between 3D points and their 2D projections.  However, in dynamic scenes this consistency breaks down, as dynamic motion violates the static assumptions underlying traditional multi-view geometry~\cite{hartley2003multiple,feng2025st4rtrack}.  Consequently, geometry reconstruction and motion estimation are often addressed separately: the former through Structure from Motion~\cite{agarwal2011building,schonberger2016structure} and Simultaneous Localization and Mapping~\cite{durrant2006simultaneous}, which assume static environments and ignore modeling dynamic content; and the latter through scene flow estimation~\cite{hadfield2011kinecting,hornacek2014sphereflow,quiroga2014dense,teed2021raft} or 3D point tracking~\cite{zhang2025tapip3d,xiao2024spatialtracker}, leveraging estimated or sensed depth maps to provide 3D information for video inputs.

Feed-forward frameworks~\cite{wang2024dust3r,leroy2024grounding,wang2024vggsfm,wang2025vggt,keetha2025mapanything} have recently emerged as effective solutions for static 3D geometry reconstruction.  These models bypass explicit triangulation by directly predicting 3D point positions from multi-view images, typically centered at a single camera's coordinate frame to handle varying camera poses.  The same architectures have been extended to dynamic scene reconstruction~\cite{jin2024stereo4d,feng2025st4rtrack,yao2025uni4d,sucar2025dynamic,lei2025mosca,han2025d}, estimating 3D positions of every observed points across time to jointly model geometry and motion.  While conceptually simple, such direct extensions are suboptimal:  predicting future or past 3D point locations in another camera's coordinate frame needs to jointly account for both global camera motion and local object motion, which is ambiguous, and therefore increasing the complexity of dynamic scene reconstruction task.

To address this limitation, our key insight is to disentangle camera motion from object motion in dynamic scenes. We make three observations: (1) motion estimation becomes simpler when camera motion is excluded, (2) accurately estimating object motion helps identify static points, and (3) camera motion naturally emerges from the multi-view geometry of these static points.  Building on these insights, we focus solely on modeling object motion by predicting 3D point positions across time within each input camera’s coordinate frame.  This formulation produces motion patterns that distinctly separate static and dynamic points, enabling robust camera motion estimation from static-point geometry predicted by existing feed-forward models. By defining motion in this disentangled space, we simplify dynamic scene reconstruction and improve the accuracy and coherence of its constituent tasks.

We introduce \methodfull{} (\method{}), a multi-task cascade feed-forward model~\cite{dai2016instance} that bootstraps motion estimation from geometry reconstruction for dynamic scene reconstruction (see Fig.~\ref{fig:teaser}). Built on the DUSt3R~\cite{wang2024dust3r} architecture, \method{} first predicts 3D point positions from two input video frames and then estimates their 3D motion as positional offsets. To enhance motion estimation, \method{} modulates attention~\cite{vaswani2017attention} between visual tokens based on predicted 3D locations. By ignoring global camera motion, \method{} predicts per-frame 3D point motion within each camera coordinate frame, producing zero motion for static points. To address the sparsity of real-world point motion annotations, we introduce a weakly-supervised loss that regularizes unlabeled point motion based on labeled point positions from the other time step. Together, these designs enable \method{} to reconstruct dynamic scenes more accurately and coherently.  

We extensively evaluate \method{} under multiple established setups.  Following~\cite{sucar2025dynamic}, we evaluate the effectiveness of dynamic scene reconstruction on Kubric-F~\cite{greff2022kubric} and Waymo~\cite{sun2020scalability} dataset.  Our \method{} outperforms previous state-of-the-arts in terms of estimating object flow on both datasets.  Meanwhile, we evaluate the effectiveness of geometry reconstruction on 5 datasets: Sintel~\cite{Butler2012ECCV}, Point Odyssey~\cite{zheng2023point}, Bonn~\cite{palazzolo2019refusion}, Kubric-F~\cite{greff2022kubric}, and KITTI~\cite{Geiger2013IJRR}.  Our \method{} achieves competitive performance than previous methods in terms of two-view depth estimation.  Our results demonstrate that our unified framework for geometry reconstruction and motion estimation does not compromise the performance of either task.  Instead, our cascade design enhances motion estimation with geometry reconstruction.  We believe this technical insight would facilitate the development of more advanced feed-forward model architecture for dynamic scene reconstruction.

\section{Preliminary}

\begin{figure}[!htb]
    \centering
    \includegraphics[width=0.82\linewidth]{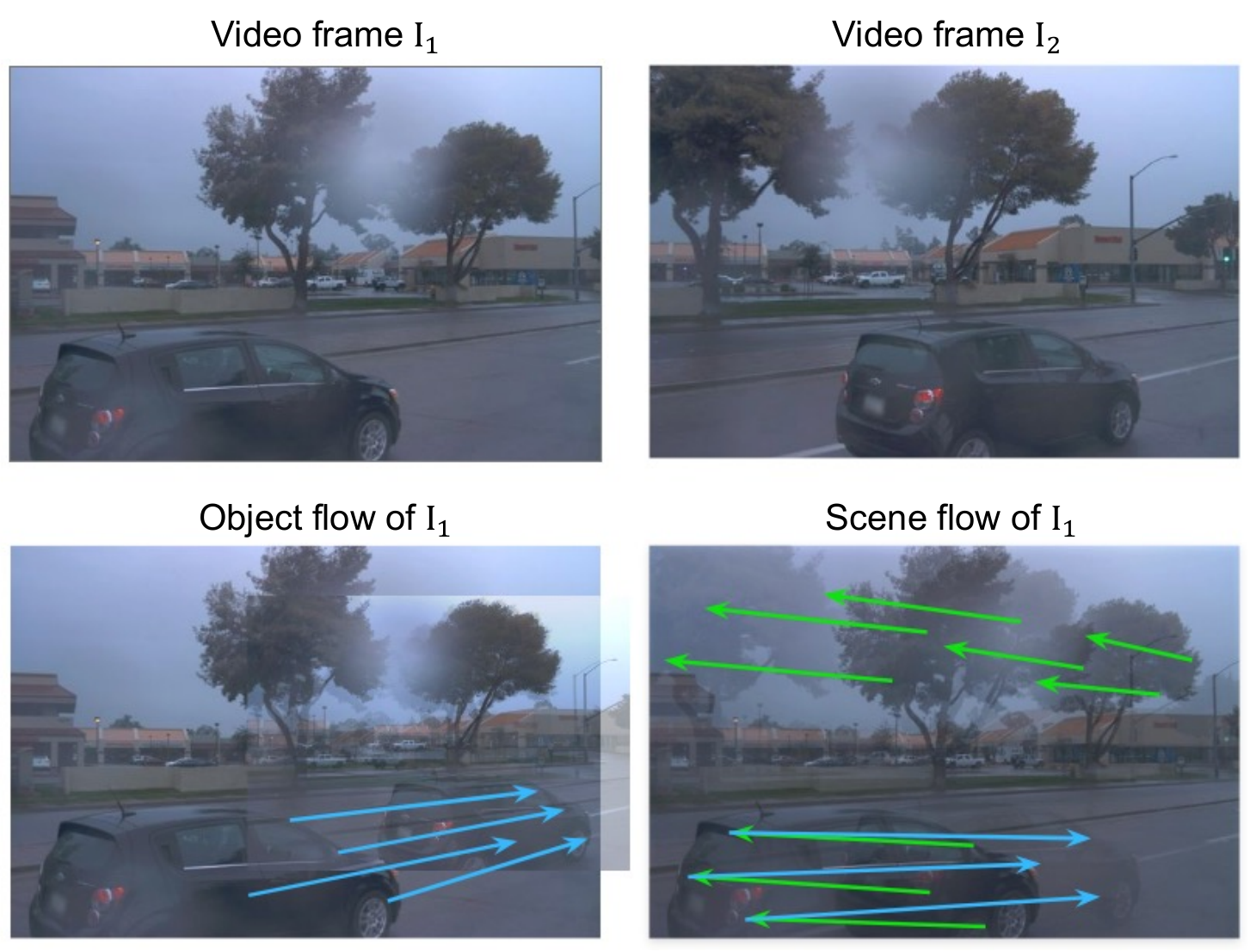}
    \vspace{4pt}
    \caption{
    \textbf{Visualization of object flow and scene flow.}
    Given two video frames $I_1$ and $I_2$, we compare two motion representations: object flow and scene flow.
    The bottom-left panel shows the object flow of frame $I_1$, where {\color{cyan}blue} arrows indicate object motion.
    The bottom-right panel shows the scene flow of $I_1$, which additionally includes camera motion, indicated by {\color{green}green} arrows.
    Object flow is less ambiguous than scene flow.
    }
    \label{fig:motion_compounding}
\end{figure}

Dynamic scene reconstruction involves reconstructing 3D scene geometry and estimating motion from RGB videos.  In this section, we briefly introduce the definition of these tasks and review the seminal feed-forward model--DUSt3R~\cite{wang2024dust3r}.

\subsection{3D Geometry Reconstruction}

The goal of 3D geometry reconstruction is to recover scene structure and estimate camera parameters from a set of input images. Following~\cite{wang2024dust3r}, we represent scene structure using a \textit{point map}--the 3D positions of all image pixels. Let $I \in \mathbb{R}^{H \times W}$ denote an RGB image with height $H$ and width $W$, and $P(\pi) \in \mathbb{R}^{H \times W \times 3}$ denote its per-pixel 3D point map expressed in coordinate frame $\pi$. This formulation allows flexible representation of 3D points in different coordinate frames. In practice, point maps are expressed in camera coordinate frames defined by the corresponding camera poses, which describe transformations between the world frame and the camera frame.

\paragraph{Feed-forward geometry reconstruction model.}
DUSt3R is a transformer-based~\cite{vaswani2017attention} feed-forward model that takes as input two images, $I_1$ and $I_2$, with corresponding camera poses $\pi_1$ and $\pi_2$. It predicts point maps $P_1(\pi_1)$ and $P_2(\pi_1)$--the 3D positions of pixels in both images, expressed in $I_1$’s camera frame. When the input order is reversed, DUSt3R produces $P_2(\pi_2)$ and $P_1(\pi_2)$, expressed in $I_2$’s camera frame.  This dual prediction enables correspondence between the same scene points expressed in different frames, such as $P_1(\pi_1)$ vs. $P_1(\pi_2)$ or $P_2(\pi_2)$ vs. $P_2(\pi_1)$.  The relative camera pose can be directly estimated via Procrustes alignment~\cite{luo1999procrustes} or the Perspective-n-Point (PnP) algorithm~\cite{lepetit2009ep}.

\subsection{3D Motion Estimation}

The goal of motion estimation is to measure the 3D point motion from an input RGB video.  Following prior work~\cite{jin2024stereo4d,sucar2025dynamic}, we define point motion as the residual between pixel-wise 3D positions across time.  Let $P_n(t, \pi)$ denote the 3D positions of pixels from the $n$-th video frame, as observed at time $t$ and expressed in coordinate frame $\pi$.  This formulation generalizes that of static-scene reconstruction, where only the current frame is considered ($n = t$), to also describe future and past point maps when $n < t$ and $n > t$.  Therefore, we refer to $P_n(t, \pi)$ as \textit{dynamic point maps}~\cite{sucar2025dynamic}.

Let $I_1$ and $I_2$ denote two video frames captured at times $t_1$ and $t_2$, with corresponding camera poses $\pi_1$ and $\pi_2$.  The 3D motion of pixels from $I_1$ can be defined in three ways, depending on how camera motion is included.  The first is \textit{scene flow}: $P_1(t_2, \pi_2) - P_1(t_1, \pi_1)$, which captures both object and camera motion.  The second is \textit{cross-frame object flow}: $P_1(t_2, \pi_2) - P_1(t_1, \pi_2)$, which describes object motion expressed in the second camera’s coordinate frame, and thus requires inferring the relative camera pose.  The last is \textit{object flow}: $P_1(t_2, \pi_1) - P_1(t_1, \pi_1)$, which ignores camera motion entirely and represents object motion in the first camera’s coordinate frame.  Among these, object flow is the least ambiguous representation, since static points yield zero motion values by construction. \fig{motion_compounding} illustrates the difference between object flow and scene flow.

\paragraph{Feed-forward dynamic scene reconstruction model.}
Existing methods~\cite{jin2024stereo4d,sucar2025dynamic,feng2025st4rtrack,han2025d} extend the DUSt3R architecture for dynamic scene reconstruction by adding independent decoders and dense prediction (DPT) heads~\cite{ranftl2021vision}. Given two input frames, $I_1$ and $I_2$, these models predict per-frame dynamic point maps--$P_1(t_1, \pi_1)$, $P_1(t_2, \pi_1)$, $P_2(t_2, \pi_1)$, and $P_2(t_1, \pi_1)$--all expressed in the first camera’s coordinate frame, resulting in a mixture of cross-frame object flow and object flow representation. 

However, such direct extensions are suboptimal. To obtain cross-frame object flow ($P_2(t_1, \pi_1) - P_2(t_2, \pi_1)$) for image $I_2$, these models must implicitly infer the relative camera pose between frames.  Moreover, future and past point maps ($P_1(t_2, \pi_1)$ and $P_2(t_1, \pi_1)$) are predicted independently from the current ones, which leads to spatio-temporal inconsistencies in dynamic scene reconstruction.

\section{Method}
\label{sec:method}

\begin{figure*}
    \centering
    \includegraphics[width=\linewidth]{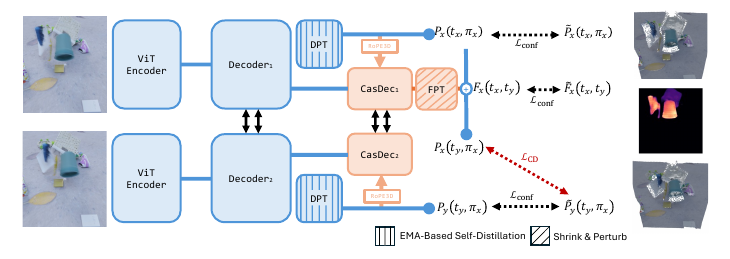}
    \vspace{10pt}
    \caption{\textbf{Overview of the framework.}  \method{} is a multi-task cascade feed-forward model for dynamic scene reconstruction.  It first performs geometry reconstruction using a DUSt3R backbone ({\color{blue} blue} modules), then feeds the predicted 3D geometry into a transformer-based decoder followed by a DPT head to estimate motion ({\color{red} red} modules).  \method{} trains the network with a fully-supervised regression loss $\mathcal{L}_{\textsf{conf}}$ based on ground-truth dynamic point maps and object flow; at the same time, it incorporates a weakly-supervised Chamfer loss {\color{red} $\mathcal{L}_{\textsf{CD}}$} that regularizes unlabeled point motion based on labeled point positions from the other time step, to handle the sparsity of motion labels.}
    \label{fig:methodology}
\end{figure*}


\method{} is a multi-task cascade feed-forward model for dynamic scene reconstruction from RGB videos. As illustrated in Fig.~\ref{fig:methodology}, it adopts a geometry-first cascade: dense 3D geometry is reconstructed first and then used to bootstrap motion estimation. To reduce ambiguity from camera motion, \method{} represents dynamics with object flow rather than motion in a camera-entangled coordinate frame. We further introduce a weakly supervised loss that regularizes unlabeled point motions, enabling the model to better leverage sparsely annotated real-world data.

\subsection{Multi-task Cascade Feed-Forward Model}
\method{} addresses geometry reconstruction and motion estimation within a cascaded multi-task framework. It first performs geometry reconstruction using a \duster backbone, then feeds the predicted 3D geometry into a transformer-based decoder followed by a DPT head to estimate motion.


For geometry reconstruction, the \duster backbone takes as input an ordered pair of video frames $I_1$ and $I_2$ with camera pose $\pi_1$ and $\pi_2$, captured at time $t_1$ and $t_2$.  It extracts visual tokens $G_1$ and $G_2$ from input images with ViT encoder~\cite{dosovitskiy2020image}, contextualizes these tokens via cross attention, and projects them to point map $P_1(t_1, \pi_1)$ and $P_2(t_2, \pi_1)$ observed at the corresponding time step while expressed in the $I_1$'s camera frame.

For motion estimation, we feed the updated tokens $G_1$ and $G_2$ into a transformer decoder, where they cross-attend to enhance spatio-temporal feature representations. Because motion estimation depends on precise 3D localization of each pixel, we inject 3D positional information into the decoder. Specifically, since each token corresponds to an image patch, we compute its 3D position by averaging the predicted 3D coordinates of pixels within that patch.  Following prior work in visual correspondence~\cite{li2022lepard} and robotic manipulation~\cite{ke20243d}, we incorporate 3D information by modulating the attention mechanism--encouraging tokens to attend more strongly to spatially nearby ones in 3D. We implement this via rotary positional embeddings~\cite{su2024roformer}, encoding relative 3D positions directly in the attention layers (see~\cite{su2024roformer} for details).

Finally, \method{} predicts the object flow $P_1(t_2, \pi_1) - P_1(t_1, \pi_1)$ of $I_1$’s pixels, describing their motion from time $t_1$ to $t_2$, by projecting token $G_1$ through a DPT head.  The token $G_1$ has been progressively updated through multiple transformer decoder layers.  To obtain the reverse object flow of $I_2$’s pixels: $P_2(t_1, \pi_2) - P_2(t_2, \pi_2)$, we simply swap the input frame order and perform another feed-forward pass.  For brevity, we denote the predicted object flows of $I_1$ and $I_2$ as $F_1(t_1, t_2)$ and $F_2(t_2, t_1)$, respectively.

\subsection{Training Objective}
For network optimization, \method{} incorporates two loss functions: a fully supervised regression loss and a weakly supervised Chamfer loss. Together, these objectives jointly supervise both the predicted point maps and object flows. In the following, we describe these loss functions using an ordered input pair $(I_1, I_2)$; the extension to the reversed input order is straightforward.

\paragraph{Scale normalization.}
Similar to \duster, \method{} produces relative, not metric, point map and object flows.  Normalizing the scale of these outputs is needed during training and inference.
We enforce spatio-temporal consistent scales of predicted point maps and object flow by scaling them with the same factor $g$ during training.  This scaling factor is:
\begin{align}
    g=\frac{\mathrm{norm}(\tilde{P}_1(t_1, \pi_1), \tilde{P}_2(t_1, \pi_1))}{\mathrm{norm}(P_1(t_1, \pi_1), P_2(t_1, \pi_1))}
\end{align}
where $\mathrm{norm}(X, Y)$ denotes the average distance between all points from point map $X$ and $Y$ and their center, while $\tilde{X}$ is the ground truth of point map $X$.

\paragraph{Fully-supervised regression loss.}
We adopt the same loss function as \duster~\cite{wang2024dust3r} to regress both point map and object flow.  First, we define a pixel-wise regression loss:
\begin{align}
    \mathcal{L}_{\textsf{regr}}(X, \tilde{X}, i) = \|g^{i} X^{i} - \tilde{X}^{i}\|_2^2
\end{align}
where $X$ denotes a dense map like the point map or object flow, $\tilde{X}$ is the ground truth of $X$, and superscript $i$ refers to the $i$-th pixel in $X$.  However, the ground-truth point maps and object flow of real-world data are often sparse and noisy.  To handle the sparsity, this fully-supervised loss is only applied to labeled pixels.  To address label noise, the \duster backbone is jointly supervised to predict pixel-wise confidence score, which weights the pixel-wise regression loss $\mathcal{L}_{\textsf{regr}}$.  Let $M(X)$ denote the set of labeled pixel indices in dense map $X$, which may correspond to the point map or object flow.  The confidence-weighted regression loss is:
\begin{align}
    \mathcal{L}_{\textsf{conf}}(X, \tilde{X}) = \sum_{i \in M(X)}C^{i} \mathcal{L}_{\textsf{regr}}(X, \tilde{X}, i) - \alpha C^i
\end{align}
where $C^i$ denotes the confidence score of pixel $i$, and $\alpha$ encourages high confidence at each pixel.

\begin{figure}[!h]
    \centering
    \includegraphics[width=\linewidth]{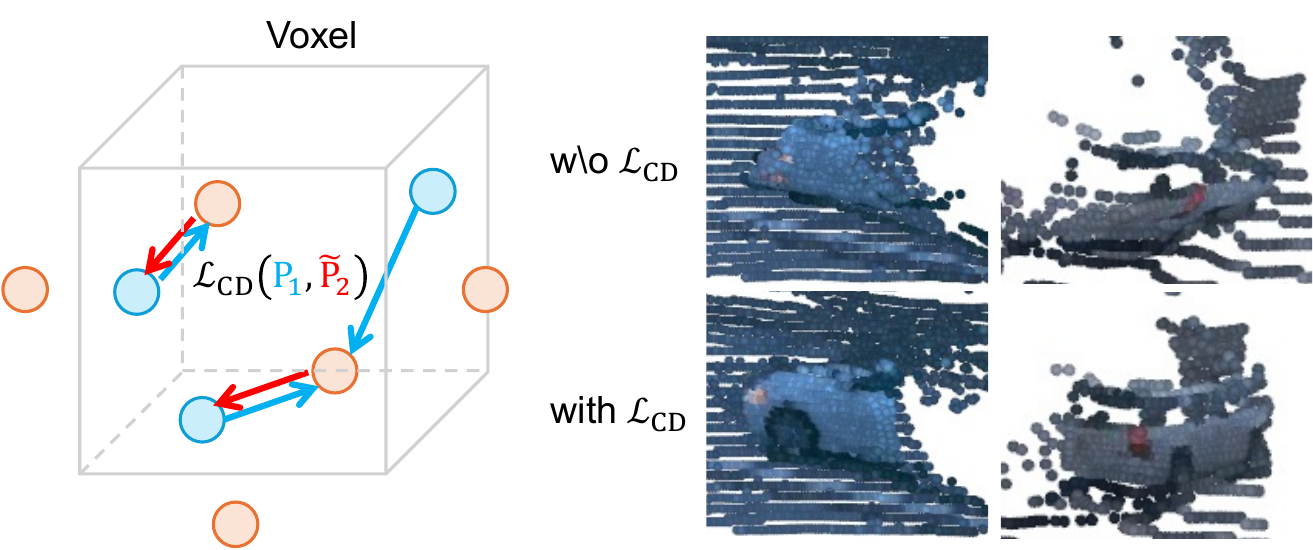}
    \vspace{10pt}
    \caption{\textbf{Weakly-supervised Chamfer loss with voxelization.}  \textbf{Left:} We propose a weakly-supervised Chamfer loss, $\mathcal{L}_{\textsf{CD}}$, to supervise a point set {\color{cyan} $P_1$} using another point set {\color{red} $\tilde{P}_2$} without requiring point correspondences. By voxelizing the 3D Euclidean space, we compute the Chamfer loss within each voxel, approximating nearest-neighbor search while filtering invalid matches. \textbf{Right:} The top row shows outputs from a model trained without the Chamfer loss, while the bottom row shows outputs when it is used. The proposed weakly-supervised loss significantly improves dynamic scene reconstruction.}
    \label{fig:chamfer_loss}
\end{figure}

\paragraph{Weakly-supervised Chamfer loss.}
Due to the sparsity of motion labels, training models with only the fully-supervised regression loss often leads to inconsistent object flow estimation within the same object.  Motivated by this observation, we propose a weakly-supervised loss that regularizes motion output of unlabeled points based on positions of labeled points from the other time steps.  In other words, we transform point positions of one frame based on the predicted object flow, and enforce them staying close to those in the other frame.  Let $S(X)$ denote the set of pixel indices in dense prediction $X$.  We define Chamfer distance $d_{\textsf{CD}}$~\cite{fan2017point} as:
\begin{align}
    d_{\textsf{CD}}(X_1, \tilde{X}_2) &= \sum_{i \in S(X_1)} \min_{j \in M(\tilde{X}_2)} \|X_1^i - \tilde{X}_2^j\|_2^2 \nonumber \\
    & + \sum_{j \in M(\tilde{X}_2)} \min_{i \in S(X_1)} \|X_1^i - \tilde{X}_2^j\|_2^2
    \label{eqn:chamfer_distance}
\end{align}
where $d_{\textsf{CD}}(X_1, \tilde{X}_2)$ equals $0$ if either $X_1$ or $\tilde{X}_2$ is empty.  Notably, calculating $d_{\textsf{CD}}$ requires searching the nearest neighbor from all points in $\tilde{X}_2$, which is computationally inefficient.  Meanwhile, due to occlusion, objects observed in one frame may be invisible in the other frame.  We propose to approximate nearest neighbor search while filtering invalid matches through voxelization.  As shown in Fig.~\ref{fig:chamfer_loss}, we discretize the 3D Euclidean space into $N$ voxels, and define the weakly-supervised loss $\mathcal{L}_{\textsf{CD}}$ as the accumulation of Chamfer distance across all voxels:
\begin{align}
    \mathcal{L}_{\textsf{CD}}(X_1, \tilde{X}_2) = \sum_{i=1}^{N} d_{\textsf{CD}}(V(X_1, i), V(\tilde{X}_2, i))
\end{align}
where $V(X, i)$ denotes the set of points within the $i$-th voxel, gathered from dense map $X$.

\paragraph{EMA-based static geometry distillation loss.}
During dynamic fine-tuning, \method{} must learn motion prediction without forgetting the static 3D reconstruction ability inherited from pretraining. We address this with EMA-based static self-distillation \cite{tarvainen2018meanteachersbetterrole}. Specifically, we maintain an exponential moving average copy of the online model as a detached teacher and encourage the student to match the teacher's predicted 3D pointmaps on valid static pixels. The loss is applied to both point predictions in the reference view and points expressed in the other view. Restricting this distillation to static regions preserves geometric reconstruction behavior while avoiding conflict with supervised motion learning on dynamic regions.

\begin{align}
    \mathcal{L}_\textsf{SD}(X_1, X_2)
    &= M^\text{static}_1 \| X_1 - X^{EMA}_1 \|^2_2 \nonumber \\
    &\quad + M^\text{static}_2 \| X_2 - X^{EMA}_2 \|^2_2
\end{align}

\paragraph{Total loss.}
Given an ordered pair of video frames $I_1$ and $I_2$, our \method{} predicts two point maps $P_1(t_1, \pi_1), P_2(t_2, \pi_1)$, a confidence score map $C$, and an object flow map $F_1(t_1, t_2)$.
The total loss of this input pair is as follows:
\begin{align}
    \mathcal{L}_{\textsf{total}} &= \mathcal{L}_{\textsf{conf}}\left( P_1(t_1, \pi_1), \tilde{P}_1(t_1, \pi_1) \right) \nonumber \\
    &+ \mathcal{L}_{\textsf{conf}}\left( P_2(t_2, \pi_1), \tilde{P}_2(t_2, \pi_1) \right) \nonumber \\
    &+ \mathcal{L}_{\textsf{conf}}\left( F_1(t_1, t_2), \tilde{F}_1(t_1, t_2) \right) \nonumber \\
    &+ \mathcal{L}_\textsf{SD}(P_1(t_1, \pi_1), P_2(t_2, \pi_1)) \nonumber \\
    &+ \lambda \mathcal{L}_{\textsf{CD}}\left( P_1(t_1, \pi_1) + F_1(t_1, t_2), \tilde{P}_2(t_2, \pi_1) \right)
    \label{eqn:total_loss}
\end{align}
where $\lambda$ weights Chamfer loss $\mathcal{L}_{\textsf{CD}}$.  In practice, adopting this weakly-supervised loss in the early stage of training would sabotage the model performance.  It offers better supervision when predicted point map $P_1(t_1, \pi_1)$ becomes accurate.  Therefore, we adopt a sigmoid scheduler to gradually increase $\lambda$ during training. 

\paragraph{Plasticity injection on flow prediction heads.}
The motion heads in \method{} are initialized from pretrained depth heads, providing a useful geometric prior but also introducing a bias toward static pointmap prediction. To help these reused heads adapt to motion estimation, we apply shrink-and-perturb \cite{ash2020warmstartingneuralnetworktraining} during fine-tuning. At scheduled steps, the flow-head parameters are interpolated with freshly initialized weights,
\[
\theta \leftarrow \lambda \theta + (1-\lambda)\theta_{\mathrm{init}},
\]
with $\lambda$ annealed toward $1$ over training. This periodically injects a diminishing amount of plasticity into the motion heads, encouraging adaptation beyond the pretrained static reconstruction prior while preserving the advantages of depth-head initialization.

\begin{table*}[ht!]
\centering
\begin{adjustbox}{width=\linewidth}
\begin{tabular}{@{}llcccccccc@{}}
\toprule
\multirow{2}{*}{Dataset} & \multirow{2}{*}{Method} & \multicolumn{4}{c}{$L_{\textsf{rel}}$} & \multicolumn{2}{c}{Object Flow} & \multicolumn{2}{c}{Relative Pose Error}\\
\cmidrule(lr){3-6} \cmidrule(lr){7-8} \cmidrule(lr){9-10}
& & $P_1(t_1)$ & $P_2(t_1)$ & $P_1(t_2)$ & $P_2(t_2)$ & $t_1 \rightarrow t_2$ & $t_2 \rightarrow t_1$ & rot & trans \\
\midrule
\multirow{4}{*}{Kub.-F} & MonST3R~\cite{zhang2024monst3r} & 0.291 & 0.286 & 0.286 & 0.284 & 0.050 & 0.049 & 0.873 & 0.181 \\
& DPM~\cite{sucar2025dynamic} & \textbf{0.041} & 0.047 & 0.049 & \textbf{0.035} & 0.033 & 0.029 & - & - \\
& DynaDUSt3R~\cite{jin2024stereo4d} & 0.108 & 0.122 & 0.104 & 0.112 & 0.054 & 0.053 & 3.507 & 0.357\\
& \method{} (ours) & 0.045 & \textbf{0.044} & \textbf{0.045} & 0.046 & \textbf{0.018} & \textbf{0.017} & \textbf{0.193} & \textbf{0.085}\\
\midrule
\multirow{4}{*}{Waymo} & MonST3R~\cite{zhang2024monst3r} & 0.170 & 0.241 & 0.332 & 0.128 & 0.096 & 0.100 & 0.787 & 1.048\\
& DPM~\cite{sucar2025dynamic} & \textbf{0.068} & \textbf{0.065} & \textbf{0.067} & \textbf{0.065} & 0.020 & 0.020 & - & - \\
& DynaDUSt3R~\cite{jin2024stereo4d} & 0.224 & 0.192 & 0.225 & 0.187 & 0.103 & 0.096 & 2.898 & 1.145 \\
& \method{} (ours) & 0.079 & 0.078 & 0.079 & 0.078 & \textbf{0.013} & \textbf{0.015} & \textbf{0.193} & \textbf{0.168} \\
\midrule
\multirow{3}{*}{Point Odyssey} & MonST3R~\cite{zhang2024monst3r} & \textbf{0.047} & 0.183 & 0.173 & \textbf{0.048} & 0.640 & 0.626 & 0.436 & \textbf{0.024} \\
& DynaDUSt3R~\cite{jin2024stereo4d} & 0.215 & 0.162 & 0.159 & 0.214 & 0.097 & 0.094 & 0.895 & 0.040 \\
& \method{} (ours) & 0.050 & \textbf{0.049} & \textbf{0.049} & \textbf{0.048} & \textbf{0.072} & \textbf{0.074} & \textbf{0.393} & 0.053 \\
\bottomrule
\end{tabular}
\end{adjustbox}
\vspace{10pt}
\caption{\textbf{Evaluation of motion estimation}. We test on 3 dynamic scene datasets: Kubric-F~\cite{greff2021kubric}, Waymo~\cite{sun2020scalability} and Point Odyssey~\cite{zheng2023point}.  We adopt three metrics: end-point error of object flow, relative point cloud err $\mathcal{L}_{\textsf{rel}}$, and relative pose transformation. \method{} matches top performance in dynamic point map estimation, and outperforms state of the art methods in other metrics.}
\label{tab:result_pose}
\end{table*}

\begin{table*}[ht!]
\centering
\begin{adjustbox}{width=\textwidth}
\begin{tabular}{@{}lcccccccccc@{}}
\toprule
& \multicolumn{2}{c}{Sintel} & \multicolumn{2}{c}{Point Odyssey} & \multicolumn{2}{c}{Bonn} & \multicolumn{2}{c}{Kubric} & \multicolumn{2}{c}{KITTI} \\
\cmidrule(lr){2-3} \cmidrule(lr){4-5} \cmidrule(lr){6-7} \cmidrule(lr){8-9} \cmidrule(lr){10-11}
& Abs Rel & $\sigma < 1.25$ & Abs Rel & $\sigma < 1.25$ & Abs Rel & $\sigma < 1.25$ & Abs Rel & $\sigma < 1.25$ & Abs Rel & $\sigma < 1.25$ \\
\midrule
MonST3R~\cite{zhang2024monst3r} & 0.347 & \textbf{0.573} & 0.065 & 0.953 & \textbf{0.071} & \textbf{0.941} & 0.166 & 0.779 & 0.069 & 0.945 \\
DPM~\cite{sucar2025dynamic} & \textbf{0.321} & 0.564 & \textbf{0.059} & \textbf{0.957} & 0.082 & 0.919 & 0.078 & 0.950 & \textbf{0.052} & \textbf{0.968} \\
\method{} (ours) & 0.367 & 0.482 &  0.064 &  0.937 &  0.082 &  0.937 &  \textbf{0.044} &  \textbf{0.963} & 0.089 & 0.951 \\ 
\bottomrule
\end{tabular}
\end{adjustbox}
\vspace{10pt}
\caption{\textbf{Depth Evaluation from 2-View Input}. We test on 5 established datasets: Sintel~\cite{Butler2012ECCV}, Point Odyssey~\cite{zheng2023point}, Bonn~\cite{palazzolo2019iros}, Kubric~\cite{greff2021kubric}, and
KITTI~\cite{Geiger2013IJRR}. Our method maintains comparable performance on depth estimation meanwhile excels on dynamic motion estimation.}
\label{tab:result_depth}
\end{table*}

\section{Experiments}
\label{sec:result}

\subsection{Experimental Setup}

\paragraph{Training dataset.}
We train \method{} on a diverse mixture of static (WildRGBD~\cite{xia2024rgbd}, BlendedMVS~\cite{yao2020blendedmvslargescaledatasetgeneralized}) and dynamic (Waymo~\cite{sun2020scalability}, Kubric-F~\cite{greff2022kubric}, Spring~\cite{mehl2023spring} and Point Odyssey~\cite{zheng2023point}) datasets. This collection provides a wide range of real-world, synthetic, indoor, and outdoor scenes. 
We closely follow prior work~\cite{wang2024dust3r,sucar2025dynamic} to prepare ground-truth point position and motion annotations.  For Point Odyssey, we annotate 2D pixels using the corresponding 3D point trajectories provided by the simulator.  For Waymo and Kubric-F, motion labels are derived from the 3D bounding box poses of each object, while 3D position labels are obtained from Lidar measurements in Waymo and from unprojected rendered depth maps in Kubric-F. Spring provides ground-truth disparity and flow in adjacent frames. We extend these labels to aggregate for long-time difference flow with consistency filtering to avoid mismatches. For WildRGBD and BlendedMVS, position labels are derived from unprojected depth maps, and motion labels are set to zero.

\paragraph{Testing dataset.}
We evaluate \method{} under the setup of motion estimation and geometry reconstruction.

For motion estimation, we consider three dynamic scene datasets--Waymo, Kubric-F and Point Odyssey, where models are evaluated on a held-out testing set.  Following ~\cite{sucar2025dynamic}, we adopt three metrics to evaluate the model performance, including (1) end-point error (EPE) of object flow, that measures the average Euclidean distance between the predicted and ground-truth object flow, (2) relative point cloud error $L_{\textsf{rel}}$, that measures the relative L2 error of the predicted and ground-truth point map, and (3) relative pose transformation, that calculates the spatial transformation needed to align the predicted and ground-truth point map. We calculate the relative pose via PnP over point maps $P_1(\pi_1), P_2(\pi_1)$.

For geometry reconstruction, we extensively evaluate on the held-out testing set of 5 datasets: Sintel~\cite{Butler2012ECCV}, Point Odyssey~\cite{zheng2023point}, Bonn~\cite{palazzolo2019refusion}, Kubric-F~\cite{greff2022kubric}, and KITTI~\cite{Geiger2013IJRR}.  Depth estimation is typically considered as the proxy task for evaluating the performance of geometry reconstruction.  We adopt two established depth metrics: absolute relative error (Abs Rel) and $\delta < 1.25$ accuracy~\cite{yang2024depth}.  See the Appendix for detailed formulation of each metric.

\paragraph{Baseline methods.}
We consider the following baseline methods: (1) MonST3R~\cite{zhang2024monst3r}, that directly extends DUSt3R to dynamic scenes.  It predicts two point maps $P_1(t_1, \pi_1)$ and $P_2(t_2, \pi_1)$, associates pixels between image $I_1$ and $I_2$ through estimated optical flow, and calculates object flow by subtracting 3D positions of associated pixels across time.  (2) DPM~\cite{sucar2025dynamic}, that learns to predict dynamic point map $P_1(t_1, \pi_1), P_1(t_2, \pi_1), P_2(t_2, \pi_1), P_2(t_1, \pi_1)$ and estimates motion by subtracting a pair of point maps.  (3) DynaDUSt3R~\cite{jin2024stereo4d}, that learns to predict object flow $P_1(t_2, \pi_1) - P_1(t_2, \pi_1)$ of $I_1$'s pixels from $t_1$ to $t_2$, and cross-frame object flow $P_2(t_1, \pi_1) - P_2(t_1, \pi_1)$ of $I_2$'s pixels reversely from $t_2$ to $t_1$.  For DPM, we report their results based on the paper as there is no publicly available code.  For DynaDUSt3R, we use its publicly released code base\footnote{\href{https://stereo4d.github.io/}{https://stereo4d.github.io/}} to train the model. We strictly follow the same hyperparameter configuration as the original paper and train a separate model for each evaluation dataset.

\paragraph{Training details.}
During training, we randomly sample image pairs within $K \in [1, 15]$ frames from a video (depending on the dataset), and resize them to a resolution randomly selected from $\{512 \times 288$, $512 \times 336$, $512 \times 384$, $512 \times 512\}$.  We train the model for $200$ epochs based on an adaptive learning rate schedule, with an initial learning $5e^{-5}$. Our model is trained on four NVIDIA ADA 6000 GPUs for 40 hours.

\subsection{Evaluating Motion Estimation}
We evaluate on the held-out test split of Waymo, Kubric-F and Point Odyssey dataset.  On Waymo, we sample $1000$ adjacent pairs of frames from unseen clips.  On Kubric, we sample $1000$ image pairs of adjacent frames from the $1470$ testing scenes.  On Point Odyssey, we sample $1000$ image pairs within $10$ frames from $109$ unseen video sections.

In this experiment, we evaluate the quality of the predicted dynamic point maps, object flows, and pose estimation. A key distinction is that all baseline methods express the 3D positions and motions of pixels from both $I_1$ and $I_2$ in the camera frame of $I_1$ ($\pi_1$).  In contrast, \method{} expresses pixels from $I_1$ and $I_2$ in their respective camera frames ($\pi_1$ and $\pi_2$). For a fair comparison, we transform each baseline’s predictions for pixels of $I_2$ into its own camera frame ($\pi_2$) using the ground-truth relative camera pose. This transformation has no effect on their quantitative metrics, as rigid spatial transformations preserve Euclidean distances.

We summarize our results in Table~\ref{tab:result_pose}. Our method \method{} demonstrates consistently strong performance across all three dynamic scene datasets. While achieving competitive results in point map estimation ($L_{rel}$) against leading methods, our approach significantly outperforms all baselines in the more challenging tasks of object flow and relative pose estimation. On the Waymo dataset, \method{} reduces object flow error by nearly 33\% improvement compared to the next-best method ($0.014$ EPE vs. $0.020$ EPE). Similarly, on Kubric-F, our model achieves a competitive state-of-the-art object flow EPE and the lowest relative pose error. This trend holds on Point Odyssey, confirming that our model's cascaded design and motion representation lead to a superior estimation of dynamic motion. We also present qualitative results in Fig.~\ref{fig:qualitative}.

\subsection{Evaluating Geometry Reconstruction}
We test on the held-out testing set of Sintel, Point Odyssey, Bonn, Kubric-F and KITTI.  The evaluation procedure strictly follows prior work~\cite{sucar2025dynamic}, considering two-view depth estimation as the proxy task.  We randomly sample pairs of images $I_1$ and $I_2$ from videos, extracting the depth map as the z-coordinates of the predicted point map of image $I_1$, and calculates Abs Rel and $\delta < 1.25$ accuracy with regard to the ground-truth depth map.  To obtain the depth map of image $I_2$, we simply swap the input order of these two images.  See~\cite{wang2024dust3r,sucar2025dynamic} for more details.

Our results are presented in Table~\ref{tab:result_depth}. Our method maintains comparable performance across datasets, proving that our cascaded design does not compromise the performance of geometry reconstruction.

\begin{table}[ht]
\centering
\begin{adjustbox}{width=0.75\linewidth}
\begin{tabular}{@{}lcccc@{}}
\toprule
\multirow{2}{*}{Dataset: Waymo} & \multicolumn{2}{c}{$L_{rel}$} & \multicolumn{2}{c}{Object Flow} \\
\cmidrule(lr){2-3} \cmidrule(lr){4-5}
& $P_1(\pi_1)$ & $P_2(\pi_2)$ & Forward & Backward \\
\midrule
\method{} (ours) & \textbf{0.075} & \textbf{0.075} & \textbf{0.006} & \textbf{0.007} \\
w/o Chamfer loss & 0.151 & 0.151 & 0.038 & 0.037 \\
w/o 3D RoPE & 0.147 & 0.146 & 0.039 & 0.038\\

\bottomrule
\end{tabular}
\end{adjustbox}
\vspace{10pt}
\caption{\textbf{Ablation studies on model components}.  We train and evaluate each component of our model on the Waymo dataset. Both the proposed weakly-supervised Chamfer loss and the injection of predicted 3D point positions into the motion estimator using RoPE~\cite{su2024roformer} are essential for strong performance.}
\label{tab:ablation}
\end{table}

\subsection{Ablation study}
We perform extensive ablation studies to assess the contribution of each model component of our \method{}.  The experiments are conducted on the Waymo dataset, where all models are trained for $200$ epochs.  We summarize the results in Table~\ref{tab:ablation} and discuss them as follows.

\paragraph{Enhancing motion estimation with geometry reconstruction.}
We validate the contribution of predicted 3D point positions to motion estimation by removing their injection into the motion estimator’s transformer decoder.  Specifically, we replace each token’s 3D position with its corresponding 2D pixel coordinates.  As in \duster, RoPE is applied to encode relative positional information in the attention layers.
This modification leads to a substantial performance drop. The model without 3D point positions exhibits $0.072$ and $0.071$ higher $L_{\textsf{rel}}$ for the predicted point maps of $I_1$ and $I_2$, respectively.  It also shows $0.033$ and $0.031$ higher EPE for the corresponding object flows.  These results confirm that the predicted 3D point positions are essential for accurate motion estimation.

\paragraph{Weakly-supervised Chamfer loss.}
We evaluate the effectiveness of the weakly-supervised loss $\mathcal{L}_{CD}$ under sparse motion annotations by removing it from the total loss (Eqn.~\ref{eqn:total_loss}). The model trained without $\mathcal{L}_{CD}$ shows a substantial degradation in performance: $L_{\textsf{rel}}$ for the predicted point maps of $I_1$ and $I_2$ increases from $0.075$ to $0.151$, and the EPE of object flow increases by $0.032$ and $0.030$, respectively.
These results indicate that the sparsity of real-world motion labels forms a major performance bottleneck, and incorporating the weakly-supervised loss helps alleviate this limitation by effectively regularizing unlabeled points.

\subsection{Qualitative analysis}
\begin{figure*}
    \centering
    \begin{adjustbox}{width=0.95\linewidth}
    \tb{@{}ccccc@{}}{0.2}{
        Input video frames & MonST3R & DynaDUSt3R & \method{} (ours) & Ground-truth \\
        \imw{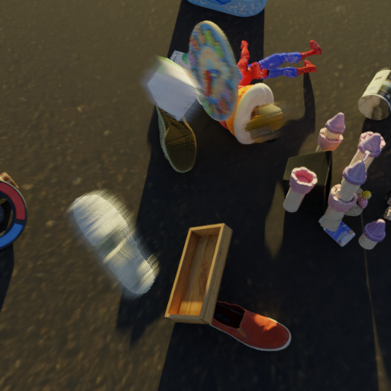}{0.19} & \imw{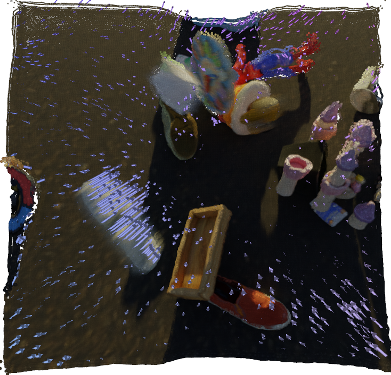}{0.19} & \imw{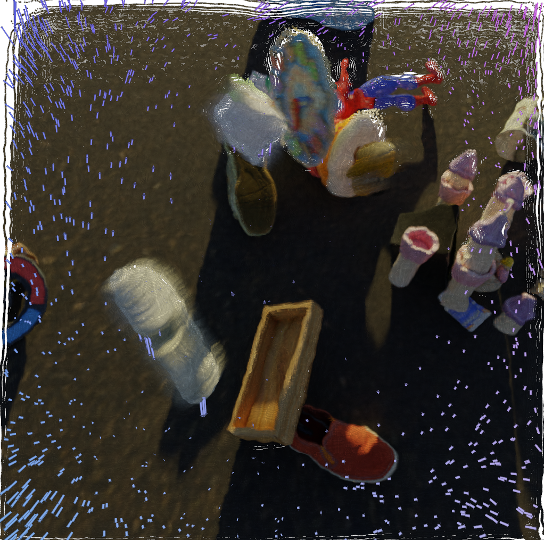}{0.19} & \imw{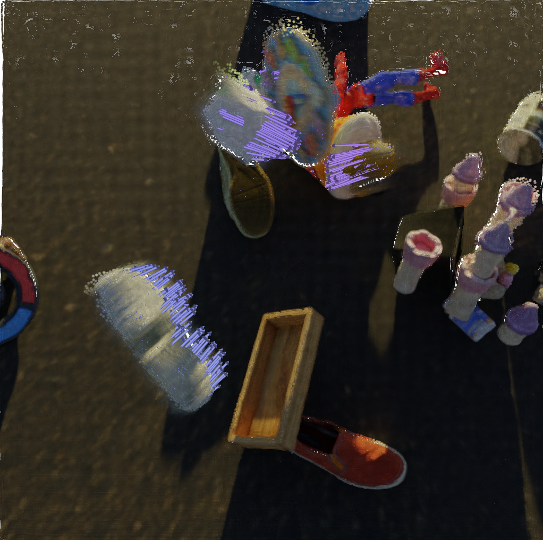}{0.19} & \imw{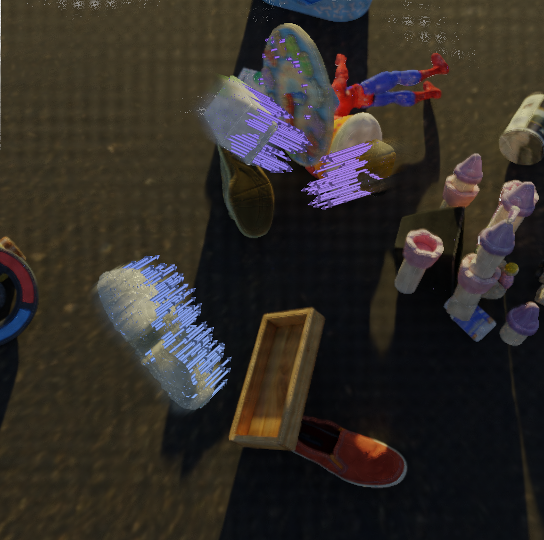}{0.19} \\ [-2pt]
        \imw{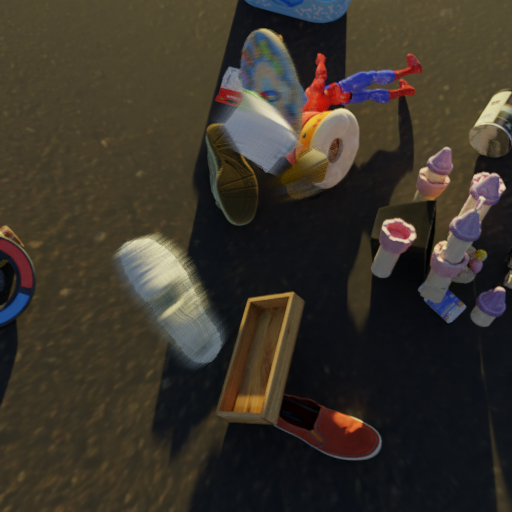}{0.19} & \imw{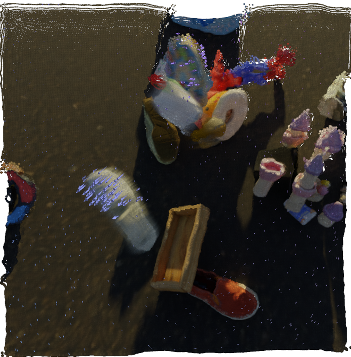}{0.19} & \imw{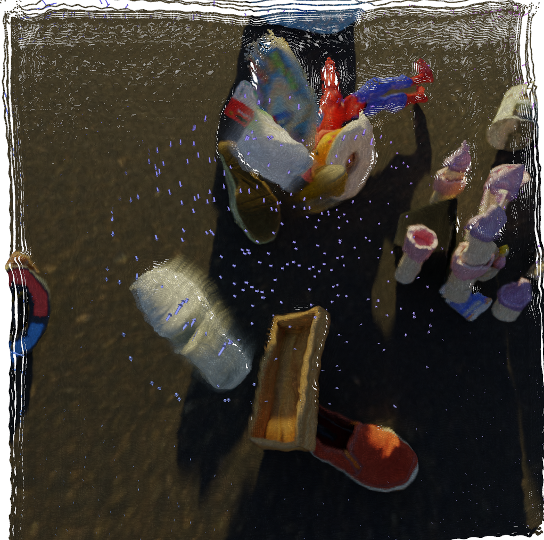}{0.19} & \imw{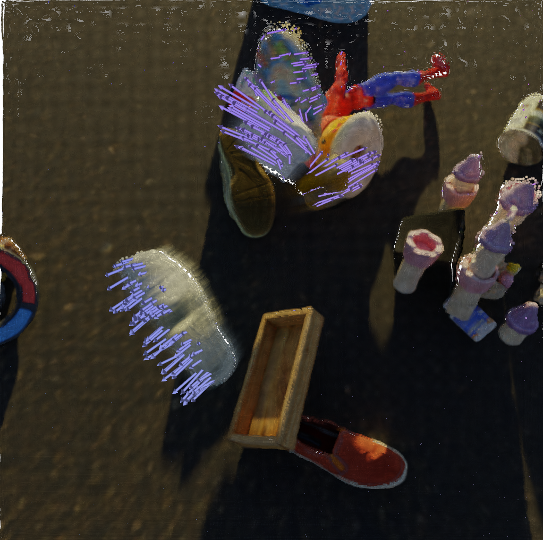}{0.19} & \imw{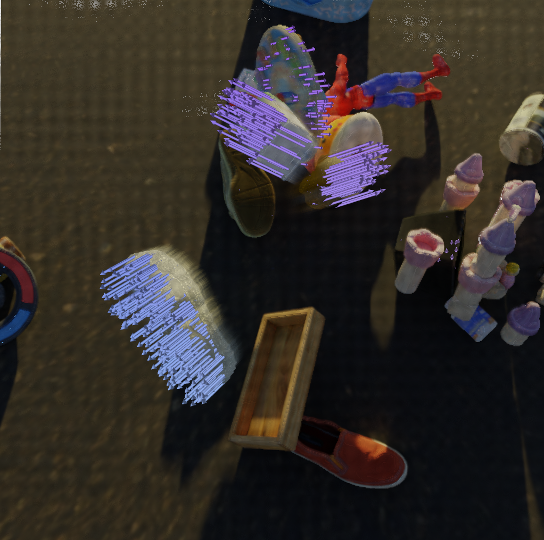}{0.19} \\
    }
    \end{adjustbox}
    \vspace{10pt}
    \caption{\textbf{Visual results of motion estimation.}  From left to right, we present the input image, object flow predicted by MonST3R~\cite{zhang2024monst3r}, DynaDUSt3R~\cite{jin2024stereo4d}, \method{}, and the ground-truth object flow.  The top and bottom row show the input image pair at two time steps.  The experiment is conducted on Kubric-F dataset.  The output motions of MonST3R and DynaDUSt3R are noisy and incorrect.  In contrast, \method{} is the best among all models for dynamic scene reconstruction.}
    \label{fig:qualitative}
\end{figure*}

\begin{figure}[h]
  \centering
  \begin{adjustbox}{width=0.7\linewidth}
    \tb{@{}ccc@{}}{0.2}{
        Input video frames & DynaDUSt3R & \method{} \\
        \imw{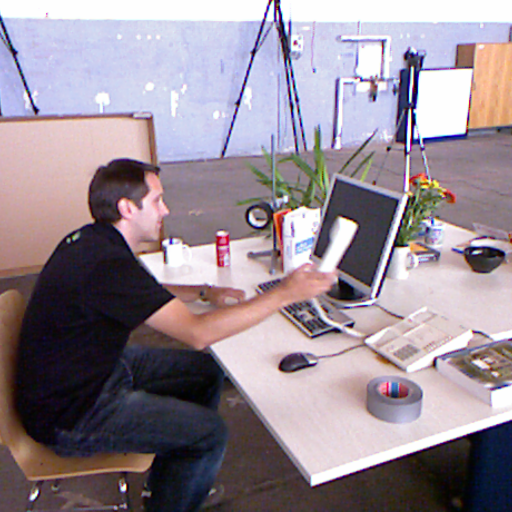}{0.3} & \imw{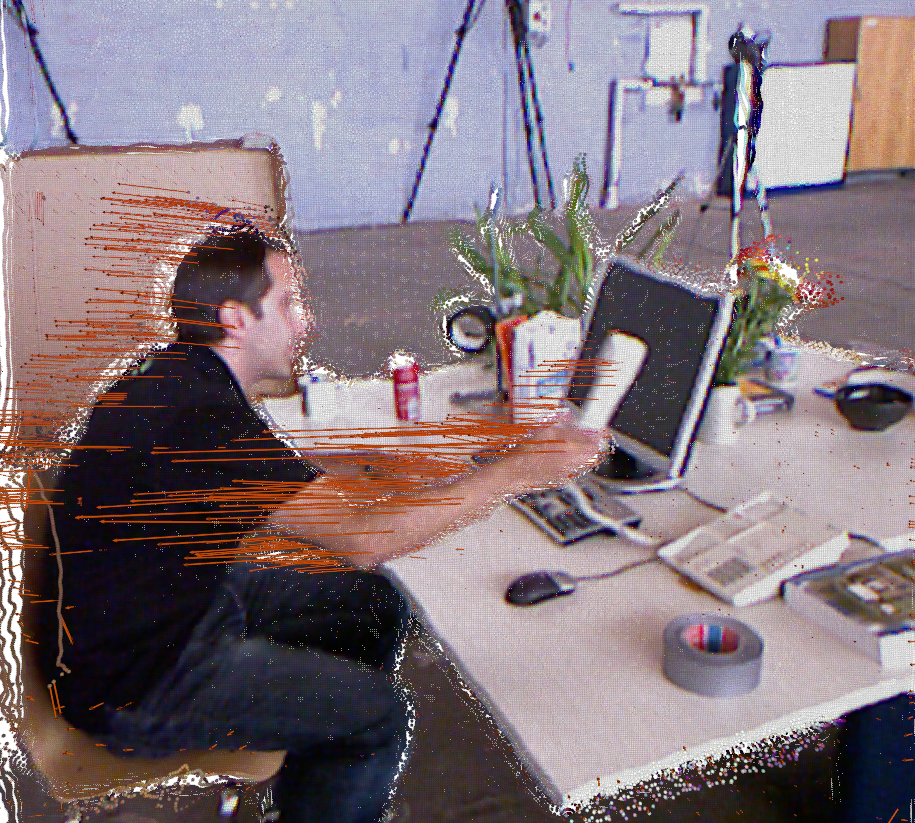}{0.3} & \imw{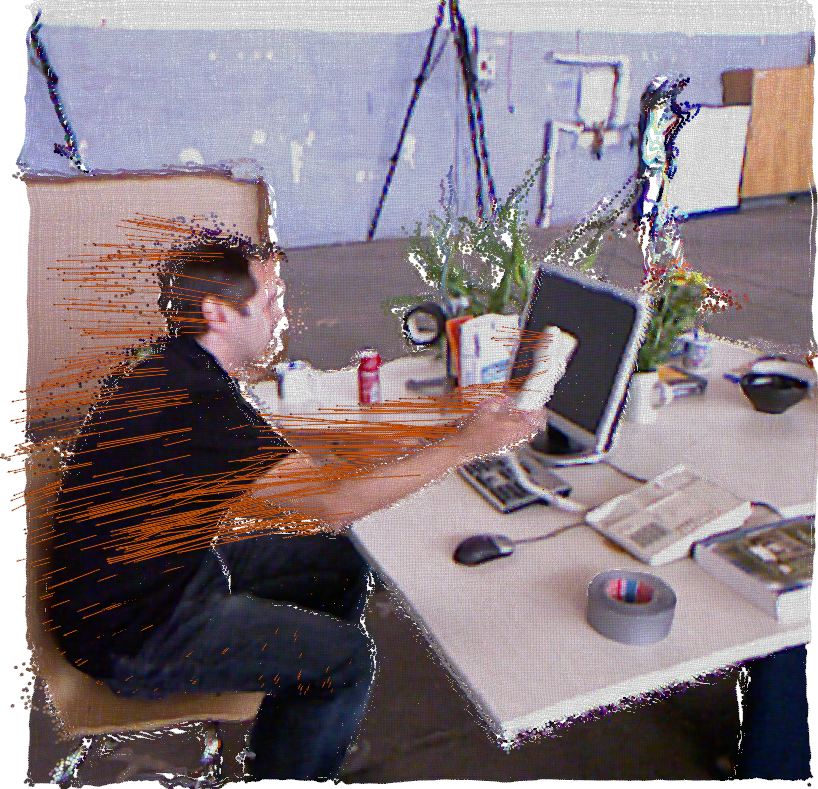}{0.3} \\ [-2pt]
        \imw{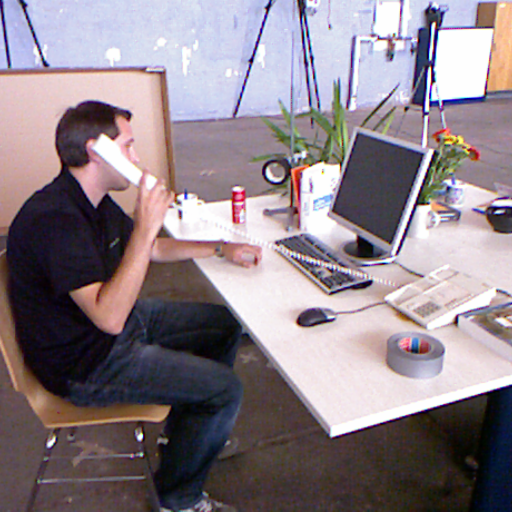}{0.3} & \imw{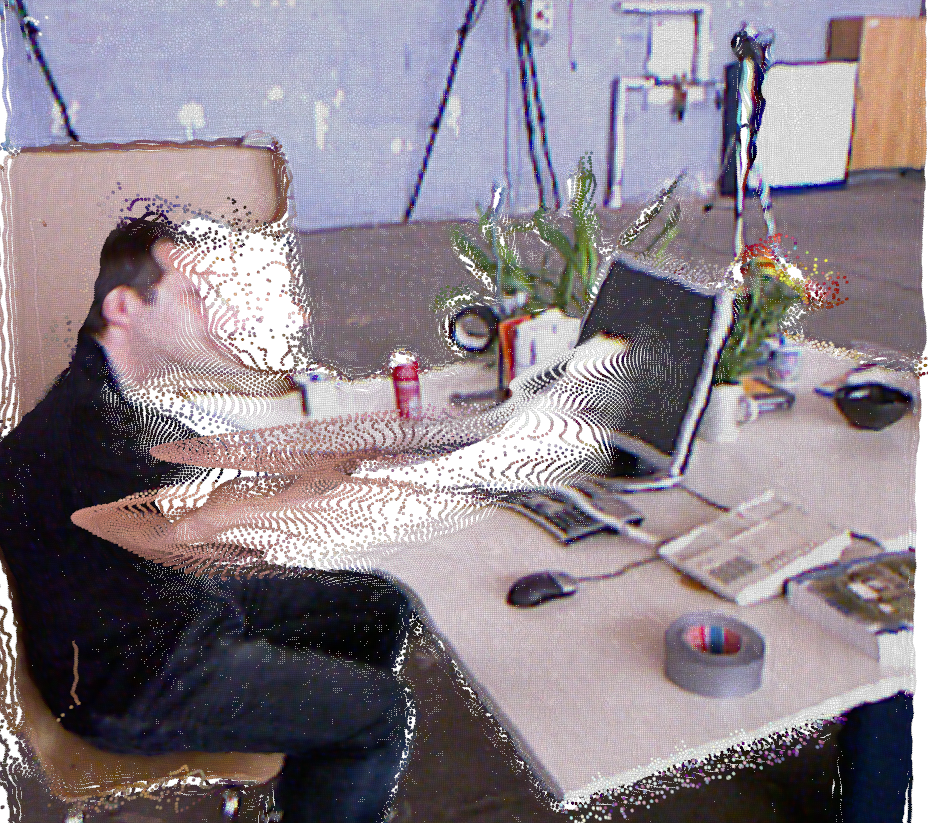}{0.3} & \imw{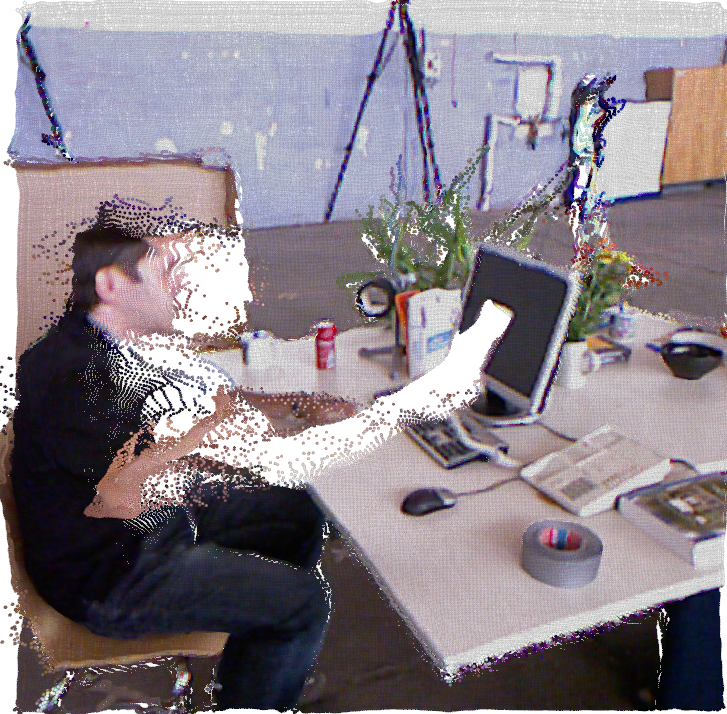}{0.3} \\
    }
    \end{adjustbox}
    \vspace{10pt}
   \caption{\textbf{Visualization on TUM RGB-D SLAM Benchmark.} The top image shows the first frame with object flow, while the bottom image shows the point map obtained by moving the first point map with the predicted object flow $P_1(t_2, \pi_1)$.}
   \vspace{-5pt}
   \label{fig:tum_rgbd}
\end{figure}

We perform qualitative analysis on selected samples from the testing sets of Kubric-F to demonstrate the effectiveness of our approach. \fig{qualitative} shows an example image pair extracted from a video clip in Kubric. MonSt3R has some success in recognizing object flows with 2D optical flow methods, but struggles on smaller objects and clearly static points in the background. DynaDUSt3R fails to detect any significant motion, which may be due to the lack of incorporating the predicted point maps during motion prediction. \method{} shows the most resemblance to the ground truth, correctly predicting several different directions for multiple objects. This shows that \method{} performs well even during scenes with complex object motion. In \fig{tum_rgbd}, we also demonstrate motion prediction performance on human hand and body movement in real-world unseen examples from TUM RGB-D SLAM \cite{sturm12iros} dataset.

\section{Related Work}



\noindent \textbf{Learning based 3D reconstruction.} The field of 3D scene reconstruction has explored diverse 3D representations, including implicit methods, meshes \cite{anciukevivcius2023renderdiffusion}, and 3D Gaussian Splatting \cite{wu20243dgs, gong2023diffpose}. Parallel to these, explicit point cloud reconstruction has seen significant advances. Foundational models like \duster\cite{wang2024dust3r} have revolutionized the field by creating a unified framework for estimating depth, camera intrinsics, and extrinsics from stereo views. Subsequent works such as MASt3R\cite{leroy2024grounding} further enable robust 3D image matching, and several other works based on \duster \cite{wang20253d, wang2025vggt, wang2025continuous} extend the model to perform global scene reconstruction from video.
Our work harnesses \duster's proven strengths in 3D scene reconstruction and alignment as a foundation, allowing our model to perform 4D sceneflow estimation with point cloud reconstructions. \\

\noindent \textbf{4D reconstruction.} Beyond static 3D, 4D reconstruction aims to model temporal dynamics. While some methods extend 3D Gaussian Splatting (3D-GS) \cite{wu20244d}, a key approach is tracking point correspondences over time. Earlier methods such as MonST3R\cite{zhang2024monst3r} combined depth with optical flow trackers, such as RAFT \cite{teed2020raft, teed2021raft}, to reconstruct 3D point movement. More recently, end-to-end networks inspired by \duster \cite{wang2024dust3r} directly predict point clouds and their motion.

A critical challenge in this area is the disentanglement of object motion from camera motion. Several works have addressed this: Monst3R \cite{zhang2024monst3r} introduced static mask prediction, integrating RAFT to enhance static background matching. DynaDUSt3R \cite{jin2024stereo4d} attempts to directly predict object motion while keeping 3D reconstruction unchanged. Dynamic Point Maps (DPM) \cite{sucar2025dynamic} address this by introducing time invariance as well as viewpoint invariance. However, by predicting motion into a single coordinate frame, these methods often struggle to fully separate object motion from camera motion. Instead, our model guarantees that object flow is predicted from the same frame as the point map. Additionally, we leverage positional information from predicted point maps to aid downstream decoder blocks in predicting displacement from the original point cloud, simplifying the motion estimation task. 
\section{Summary}
\label{sec:summary}
This paper introduces \method{}, a multi-task cascade feed-forward model for dynamic scene reconstruction from RGB videos. The model first reconstructs 3D geometry and subsequently bootstraps motion estimation from the predicted structure. By expressing 3D motion within each pixel's corresponding camera frame, \method{} effectively disentangles object motion from the ambiguity of camera movement. The framework also incorporates a weakly-supervised loss to address the common challenge of sparse motion labels in real-world data.

We evaluated \method{} on multiple dynamic-scene datasets, where it consistently outperformed all baselines for motion estimation tasks. Furthermore, it achieved competitive performance with state-of-the-art methods for geometry reconstruction. These results demonstrate that a cascaded, geometry-informed approach enhances motion estimation accuracy without compromising the performance of either task, leading to improved overall geometric reasoning.

\paragraph{Limitation \& Future work.}
Our \method{} operates on a pair of video frames and predicts the motion between them, but extending this framework to long-horizon videos remains a significant challenge. In addition, \method{} currently relies on the \duster backbone for geometry reconstruction, which can degrade in the presence of dynamic object motion. Developing methods that maintain robust geometry reconstruction in dynamic scenes is an important direction for future work. We agree that multi-frame reconstruction is an important extension. We deliberately study the two-frame setting to isolate the contribution of geometry priors to dynamic reconstruction. Importantly, our cascaded design is not tied to a specific backbone or temporal window, and can serve as a plug-and-play dynamic module for multi-frame models such as VGGT \cite{wang2025vggt} and DA3 \cite{lin2025depth3recoveringvisual}. Their temporally aggregated geometry can provide stronger correspondence priors, while our object-flow module separates camera-induced motion from independently moving regions. Moreover, the 3D RoPE formulation encodes geometry in a spatially consistent coordinate system and can be extended with temporal indexing to support correspondence reasoning across longer sequences.


\subsection*{Acknowledgments}
This work was also supported by the AMD University Program AI \& HPC Cluster. We further acknowledge Kuo-Guang Tsai for his technical support on the AMD system cluster infrastructure. This research was also supported by the National Science and Technology Council (NSTC), Taiwan, under Grants 115-2634-F-002 -012 - and 115-2634-F-002 -001 -

\bibliographystyle{plainnat}
\bibliography{references}

\clearpage
\appendix
\section{Appendix}

\subsection{Depth Evaluation Metrics}

For the depth evaluation metrics, we adopt the single-frame scale and shift-invariant evaluation protocol from Monst3r \cite{zhang2024monst3r}. We report the Abs Rel $\downarrow$ and $\delta < 1.25 \uparrow$, consistent with prior works \cite{hu2024-DepthCrafter, yang2024depth}. To enable scale-shift invariant evaluation, we adapt the optimization method from Monst3r to align the estimated depth $d$ with the ground truth depth $\hat{d}$. The evaluation metrics are defined as follows:

\begin{equation}
    \text{Absolute Relative Error (Abs Rel)}: \frac{|\hat{d} - d|}{\hat{d}}
\end{equation}

\begin{equation}
   \text{Percentage of inlier } \delta < 1.25: \max\left(\frac{d}{\hat{d}}, \frac{\hat{d}}{d}\right) < 1.25
\end{equation}

\subsection{Pose Estimation}

To retrieve the relative pose between two frames, we perform PnP RANSAC \cite{lepetit2009epnp, fischler1981random} between the reconstructed point clouds $P_i(t_i, \pi_i) \text{ and } P_i(t_i, \pi_j)$. A critical factor for accurate pose estimation is reference point selection, as static points yield more robust and precise camera motion estimates. We derive a static object mask from the confidence scores of the flow prediction head, based on the observation that confidence scores in static regions (where the flow head is expected to output zero) are significantly higher than in dynamic regions. Specifically, we generate the static mask by applying a threshold to the confidence score map; in this work, we set the threshold as $\text{conf} \ge 7$. Finally, we apply RANSAC within this masked static region to obtain the relative pose between the two frames.

\begin{figure}
    \centering
    \includegraphics[width=\linewidth]{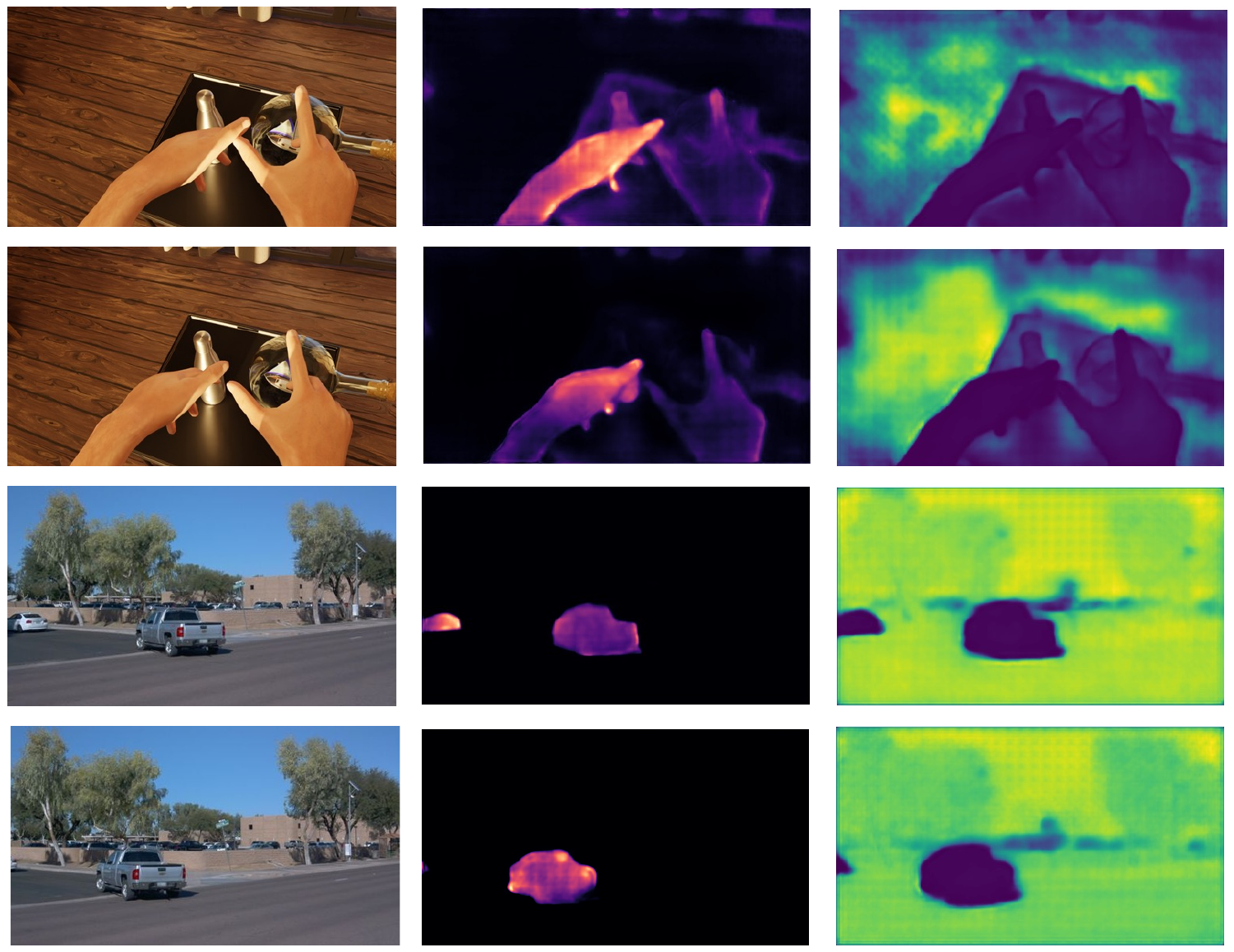}
    \caption{\textbf{Visualizations on scale of pixel-wise scene flow and static confidence map.}  \textbf{Left:} Input images. \textbf{Middle:} Visualization of the predicted object flow's $L_2$ norm. This heatmap highlights motion magnitude, allowing for a clear distinction between dynamic and static regions. \textbf{Right:} Confidence maps generated by the flow prediction head. As static areas exhibit significantly higher confidence scores compared to dynamic ones, we further leverage this property to identify stable regions for relative pose estimation. }
    \label{fig:flowconf}
\end{figure}

\subsection{Object-flow representation.}
We additionally compare object-flow and scene-flow training on PointOdyssey under identical settings. All values are EPE, with errors also reported separately for dynamic and static regions.

\vspace{-5pt}
\begin{table}[h]
\centering
\setlength{\tabcolsep}{3pt}
\scriptsize
\resizebox{\columnwidth}{!}{%
\begin{tabular}{@{}lcccc@{}}
\toprule
& Object Flow $\downarrow$ 
& Scene Flow $\downarrow$ 
& Dynamic Scene Flow $\downarrow$ 
& Static Scene Flow $\downarrow$ \\
\midrule
SM4RT (object flow) 
& \textbf{0.088} & \textbf{0.108} & \textbf{0.129} & \textbf{0.053} \\
SM4RT (scene flow)  
& 0.106 & 0.128 & 0.147 & 0.112 \\
\bottomrule
\end{tabular}%
}
\label{tab:object_scene_flow}
\vspace{-10pt}
\end{table}

Object-flow training improves all metrics, directly supporting our design. Conceptually, object flow factors out camera motion and provides a structured target: static regions should have near-zero motion, while independently moving objects retain coherent motion. This is also consistent with robotics works using object-centric 3D motion or articulation flow as action-relevant representations \cite{eisner2024flowbot3dlearning3darticulation, yuan2024generalflowfoundationaffordance}.

\subsection{Voxelization for Chamfer Distance Loss}

To balance accurate nearest neighbor search with computational efficiency, we discretize the coordinate space into grids with a resolution of 1000 per axis, resulting in $N=1000^3$ voxels. Since camera movement results in partial overlap between input point clouds, we restrict the Chamfer distance loss to be computer only on intersecting regions. These are efficiently identified via voxel occupancy intersection. Finally, we employ a sigmoid scheduler to gradually increase the weight of the Chamfer distance loss $\mathcal{L}_{\text{CD}}$ from 0 to 4.

\subsection{Pretrained Weight Setup}

Instead of training from scratch, \method{} is fine-tuned from the MASt3R \cite{mast3r_eccv24} checkpoint. We initialize the encoder, decoders, and depth prediction heads with their respective pretrained weights. Furthermore, we initialize the flow prediction head using the pretrained weights of the depth prediction head. During training, the weights of the encoder are frozen, while all other components remain trainable.

\subsection{Details about Training Datasets}
Obtaining ground truth object flow labels across arbitrary frame intervals is challenging when point trajectories in world coordinates are unavailable. For Kubric, we modify the preprocessing pipeline in TapVid3D \cite{koppula2024tapvid3d} to transform ground truth object poses into relative poses across different timestamps, thereby generating pseudo ground truth data for flow supervision. For Waymo, we derive the relative object pose by aligning the eight vertices of the ground truth bounding boxes. Finally, for Spring, we aggregated the optical flow on consecutive frames with a cycle consistency filtering on selected clips where the cameras stay still. The estimated pose is then applied to points within the corresponding bounding boxes to generate object flow data. The properties of the datasets used for training are listed in Table~\ref{tab:dataset}.

\begin{table}[ht]
\centering
\begin{adjustbox}{width=\linewidth}
\begin{tabular}{@{}lcc@{}}
\toprule
Dataset Name & Object Flow Source & Camera Movement \\
\midrule
Kubric-F \cite{greff2022kubric} & Seg. Mask + Object Pose & $\checkmark$ \\
Waymo \cite{sun2020scalability} & Object-wise Bounding Box & $\checkmark$ \\
Point Odyssey \cite{zheng2023point} & Ground Truth Point Trajectory & $\checkmark$ \\
Spring \cite{mehl2023spring} & Aggregated Optical Flow & $\times$ \\
BlendedMVS \cite{yao2020blendedmvslargescaledatasetgeneralized} & - & $\checkmark$ \\
WildRGBD \cite{xia2024rgbd} & - & $\checkmark$ \\
\bottomrule
\end{tabular}
\end{adjustbox}
\vspace{10pt}
\caption{\textbf{Training dataset overview.} We utilize various ground truth annotations to derive both ground truth and pseudo-ground truth object flow, enabling the training of our dynamic model.}
\label{tab:dataset}
\end{table}

\subsection{Experiments on extracting dynamic regions}

\begin{figure}[h!]
  \centering
  \includegraphics[width=0.7\linewidth]{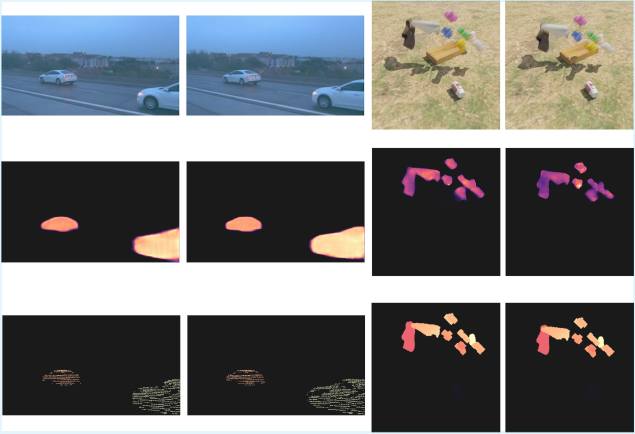}
  \vspace{10pt}
   \caption{\textbf{Visualization of the flow magnitude.} The second row shows the predicted flow with the ground-truth in the third row. }
   \label{fig:dynamic_mask}
   \vspace{-5pt}
\end{figure}

In \fig{dynamic_mask}, we visualize the predicted flow magnitudes on each image. By applying a simple threshold to the predicted flow, \method{} reliably isolates dynamic object regions, highlighting its potential for automated dynamic mask extraction.

\subsection{More visualizations}
\begin{figure*}[!h]
    \centering
    \includegraphics[
        width=\textwidth,
        keepaspectratio
    ]{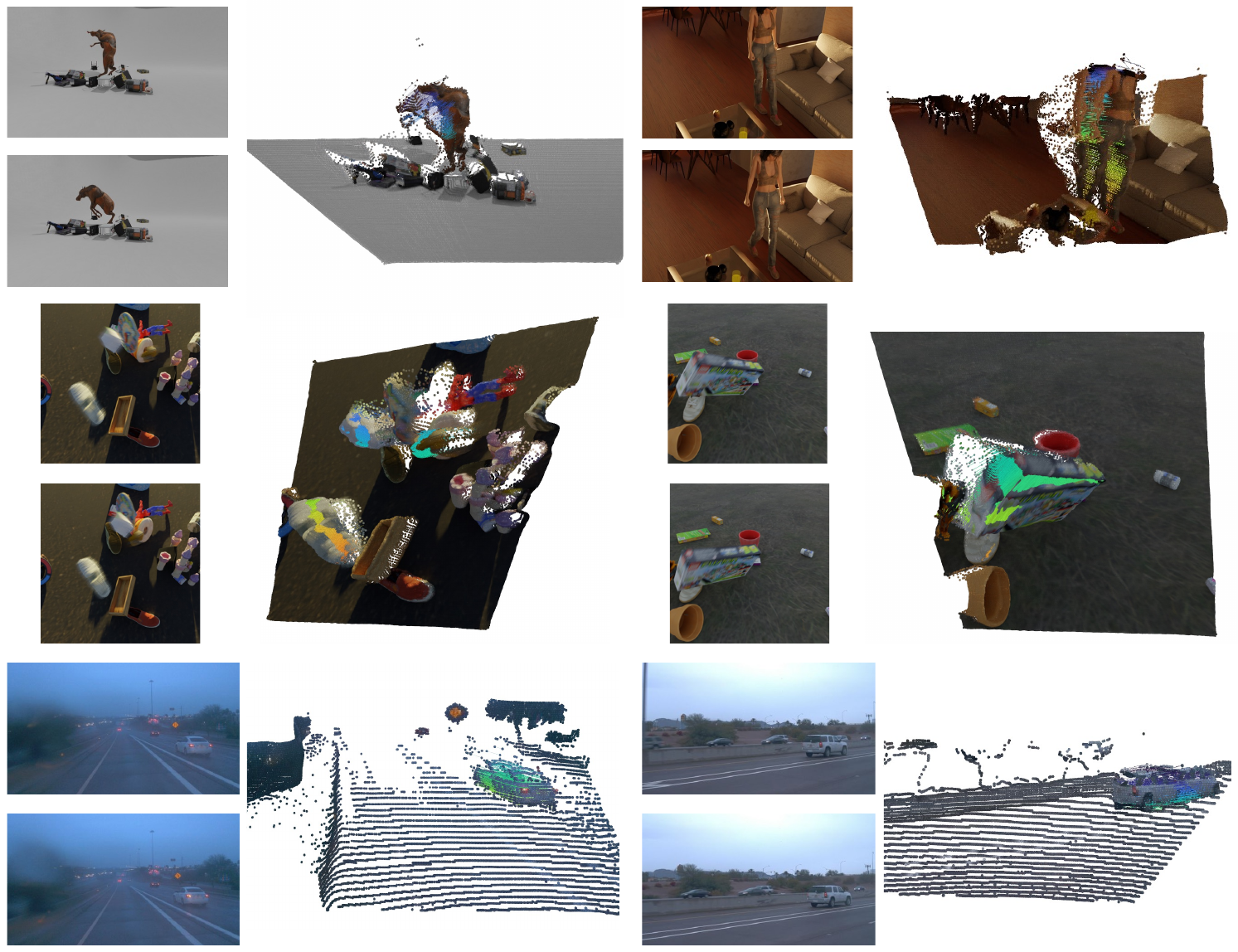}
    \vspace{4pt}
    \caption{
    \textbf{Visualizations of predicted point clouds and object flow from \method{}.}
    }
    \label{fig:vis}
    \vspace{-5pt}
\end{figure*}

We provide more visualization of \method{} on \fig{vis}.



\end{document}